\documentclass[sigconf]{acmart}
\usepackage{subcaption}
\usepackage[ruled,vlined]{algorithm2e}

\AtBeginDocument{%
  \providecommand\BibTeX{{%
    \normalfont B\kern-0.5em{\scshape i\kern-0.25em b}\kern-0.8em\TeX}}}

\begin{document}

\acmConference[KDD'20 Workshops: AIoT'20]{The 3rd International Workshop on Artificial Intelligence of Things at KDD 2020: The 26th ACM SIGKDD International Conference on Knowledge Discovery and Data Mining}{August 22-27, 2020}{San Diego, CA}
\acmBooktitle{The 3rd International Workshop on Artificial Intelligence of Things (AIoT'20) at KDD 2020: The 26th ACM SIGKDD International Conference on Knowledge Discovery and Data Mining, August 22-27, 2020, San Diego, CA}

\title{Resource-Constrained Federated Learning with Heterogeneous Labels and Models}

\author{Gautham Krishna Gudur}
\affiliation{%
  \institution{Global AI Accelerator, Ericsson}}
\email{gautham.krishna.gudur@ericsson.com}

\author{Bala Shyamala Balaji}
\affiliation{%
  \institution{Indian Institute of Technology, Madras}}
\email{balashyamala@gmail.com}

\author{Perepu Satheesh Kumar}
\affiliation{%
  \institution{Ericsson Research}}
\email{perepu.satheesh.kumar@ericsson.com}

\renewcommand{\shortauthors}{Gautham Krishna Gudur et al.}

\sloppy

\begin{abstract}
Various IoT applications demand resource-constrained machine learning mechanisms for different applications such as pervasive healthcare, activity monitoring, speech recognition, real-time computer vision, etc. This necessitates us to leverage information from multiple devices with few communication overheads. Federated Learning proves to be an extremely viable option for distributed and collaborative machine learning. Particularly, on-device federated learning is an active area of research, however, there are a variety of challenges in addressing statistical (non-IID data) and model heterogeneities. In addition, in this paper we explore a new challenge of interest -- to handle \textit{label heterogeneities} in federated learning. To this end, we propose a framework with simple $\alpha$-weighted federated aggregation of scores which leverages overlapping information gain across labels, while saving bandwidth costs in the process. Empirical evaluation on Animals-10 dataset (with 4 labels for effective elucidation of results) indicates an average deterministic accuracy increase of at least $\sim$16.7\%. We also demonstrate the on-device capabilities of our proposed framework by experimenting with federated learning and inference across different iterations on a Raspberry Pi 2, a single-board computing platform.
\end{abstract}

\ccsdesc[500]{Computing methodologies~Neural networks}
\ccsdesc[300]{Computing methodologies~Distributed computing methodologies}

\keywords{On-Device Federated Learning, Heterogeneous Labels, Heterogeneous Models, Transfer Learning}

\maketitle

\section{Introduction}
\label{section:introduction}

Contemporary machine learning, particularly deep learning has led to major breakthroughs in various domains, such as computer vision, natural language processing, speech recognition, Internet of Things (IoT), etc. Particularly, on-device machine learning has spiked up a huge interest in the research community owing to the compute capabilities vested in resource-constrained devices like mobile and wearable devices. Sensor data from various IoT devices have a vast amount of incoming data which have massive potential to leverage such on-device machine learning techniques on-the-fly to transform them into meaningful information coupled with supervised, unsupervised and/or other learning mechanisms. With the ubiquitous proliferation of such personalized IoT devices, collaborative and distributed learning is now possible more than ever to help best utilize the information learnt from multiple devices.

However, such collaborative data sharing across devices might always not be feasible owing to privacy concerns from multiple participants. Moreover, users might not prefer nor have any interest in sending their sensitive data to a remote server/cloud, particularly in fields like health-care, defense, telecommunication, etc. With the advent of \textit{Federated Learning (FL)} \cite{cite:fedavg}, \cite{cite:federated_learning}, it is now possible to effectively train a global/centralized model without compromising on sensitive data of various users by enabling the transfer of model weights and updates from local devices to the cloud, instead of conventionally transferring the sensitive data to the cloud. A server has the role of coordinating between models, however most of the work is not performed by a central entity anymore but by a federation of users. The \textit{Federated Averaging (FedAvg)} algorithm \cite{cite:fedavg} was proposed by McMahan et al. which aggregates the model parameters of each client (local device) by combining local Stochastic Gradient Descent (SGD) through a server. Federated learning has been an active and challenging area of research in  solving problems pertaining to secure communication protocols, device and statistical heterogeneities, and privacy preserving networks \cite{cite:challenges_methods}.

Federated Learning deals with various forms of heterogeneities like device, system, statistical heterogeneities, etc. \cite{cite:challenges_methods}, \cite{cite:client_selection}. Particularly, FL in IoT and edge computing scenarios, statistical heterogeneities have gained much visibility as a research problem predominantly owing to the non-IID (non-independent and identically distributed) nature of the vast amounts of streaming real-world data incoming from distinct distributions across devices. This leads to challenges in personalized federation of devices, and necessitates us to address various heterogeneities in data and learning processes for effective model aggregation.

One important step in this direction is the ability of end-users to have the choice of architecting their own models, rather than being constrained by the pre-defined architecture mandated by the global model for model aggregation. One effective way to circumvent this problem is by leveraging the concept of knowledge distillation \cite{cite:knowledge_distillation}, wherein the disparate local models distill their respective knowledge into various \textit{student models} which has a common model architecture, thereby effectively incorporating model independence and heterogeneity. This was proposed by Li et al. in FedMD \cite{cite:FedMD}. However, as much independence and heterogeneity in architecting the users' own models is ensured in their work, they do not guarantee \textit{heterogeneity and independence in labels across devices}. 

For a concrete example, a central intelligent system/service acts as the global cloud/server with multiple telecommunication operators being the clients in a typical federated learning scenario. However, most operators (clients) typically choose to have their own machine learning model architectures, and also have their own labels which are governed by few standardized and also localized company bodies. This necessitates seamless interaction of the local clients with the global cloud/server independent of model architectures and labels. Many such scenarios and settings with such heterogeneous labels and models exist in federated IoT/on-device settings, such as behaviour/health monitoring, activity tracking, keyword spotting, next-word prediction, etc. Few works address handling new labels in typical machine learning scenarios, however, to the best of our knowledge, there is no work which addresses this important problem of \textit{label and model heterogeneities} in non-IID federated learning scenarios.

The main scientific contributions in this work are as follows:
\begin{itemize}
    \item Enabling end-users to build and characterize their own preferred local architectures in a federated learning scenario, so that effective transfer learning happens between the global and local models.
    \item A framework to allow flexible heterogeneous selection of labels by showcasing scenarios with and without overlap across different user devices, thereby leveraging the information learnt across devices pertaining to those overlapped classes.
    \item Empirical demonstration of the framework's ability to handle different data distributions (statistical heterogeneities and non-IIDness) from various user devices.
    \item Demonstrating the feasibility of on-device personalized federated learning from incoming real-world data independent of users, capable of running on simple mobile and wearable devices.
\end{itemize}

\begin{figure}[ht]
  \centering
  \includegraphics[width=\linewidth]{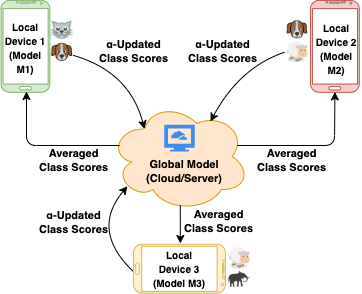}
  \caption{Overall Architecture of the proposed framework with \textit{Weighted Local Update with $\alpha$}. Each mobile device can consist of disparate set of local labels and models, and they interact with the global model (cloud/server). The models in each local device are first updated using weighted $\alpha$ score, the respective class scores are then aggregated in the global model, and the updated consensus is again distributed across local models.}
  \Description{Three different mobile devices consisting of different labels interact with a single global model which is the cloud/server.}
  \label{fig:block_diagram}
\end{figure}

\begin{table*}[ht]
\caption{Model Architectures (filters/units in each layer indicated within parentheses), Labels and Number of Images per federated learning iteration across user devices. Note the disparate model architectures and labels across users.}
\label{table:models_labels_iterations}
\begin{tabular}{|c|c|c|c|c|}
\hline
\textbf{}                     & \textbf{User\_1}                                                                   & \textbf{User\_2}                                                                       & \textbf{User\_3}                                                                & \textbf{Global\_User}        \\ \hline
\textbf{Model Architecture}   & \begin{tabular}[c]{@{}c@{}}2-Layer CNN (16, 32)\\ Softmax Activation\end{tabular} & \begin{tabular}[c]{@{}c@{}}3-Layer CNN (16, 16, 32)\\ ReLU Activation\end{tabular} & \begin{tabular}[c]{@{}c@{}}2-Layer CNN (16, 32)\\ ReLU Activation\end{tabular} & --                          \\ \hline
\textbf{Labels}               & \{Cat, Dog\}                                                                        & \{Dog, Sheep\}                                                                          & \{Sheep, Elephant\}                                                              & \{Cat, Dog, Sheep, Elephant\} \\ \hline
\textbf{Images per iteration} & \{500, 500\} = 1000                                                                 & \{500, 500\} = 1000                                                                     & \{500, 500\} = 1000                                                              & \{500, 500, 500, 500\} = 2000 \\ \hline
\end{tabular}
\end{table*}

\section{Related Work}
\label{section:related_work}

Federated Learning has contributed vividly in enabling distributed and collective machine learning across various devices. Federated learning and differentially private machine learning have/soon will emerge to become the de facto mechanisms for dealing with sensitive data, and data protected by Intellectual Property rights, GDPR, etc \cite{cite:federated_learning}. Federated Learning was first introduced by McMahan et al. in \cite{cite:fedavg}, and new challenges and open problems to be solved, aggregation and optimization methods, \cite{cite:challenges_methods}, strategies \cite{cite:federated_learning_strategies}, and multiple advancements \cite{cite:advances} have been proposed and addressed in many interesting recent works.

Particularly for Federated Learning in IoT and pervasive (mobile/wearable/edge) devices, important problems and research directions are addressed FL on mobile and edge networks in this survey \cite{cite:federated_mobile_survey}. Federated Optimization for on-device applications is discussed in \cite{cite:federated_opt}.

Multiple device/client and system heterogeneities, including client communication mechanisms and fair resource allocation, inherently making most of them optimization problems, and are addressed in various works \cite{cite:fedprox_heterogeneity}, \cite{cite:federated_heterogeneous}, \cite{cite:client_selection}, \cite{cite:fair_resource_allocation}. Personalized federated learning closely deals with optimizing the degree of personalization and contribution from various clients, thereby enabling effective aggregation as discussed in \cite{cite:personalized_fed}.

Recent works on convergence of Federated Averaging (FedAvg) algorithm on disparate data distributions -- non-IID data, and creating a small subset of data globally shared between all edge devices are proposed in \cite{cite:federated_non_iid}, \cite{cite:convergence_non_iid} respectively. Mohri et al. propose Agnostic Federated Learning \cite{cite:agnostic_federated}, which addresses about handling any target data distribution formed by a mixture of client distributions.

FedMD \cite{cite:FedMD}, which we believe to be our most closest work, deals with heterogeneities in model architectures, and addresses this problem using transfer learning and knowledge distillation \cite{cite:knowledge_distillation}, and also uses an initial public dataset across all labels (which can be accessed by any device during federated learning). On similar grounds, federated distillation and augmentation in non-IID data distributions are used in \cite{cite:fed_distill_aug}. Current federated learning approaches predominantly handle same labels across all the users and do not provide the flexibility to handle unique labels. However, in many practical applications, having unique labels for each local client/model is a very viable and common scenario owing to their dependencies and constrains on specific regions, demographics, etc. To the best of our knowledge, none of the works take into account, label and model heterogeneities.

The rest of the paper is organized as follows. Section \ref{section:our_approach} discusses the problem formulation of handling heterogeneous labels and models in on-device federated learning scenarios (section \ref{section:problem_formulation}), and section \ref{section:proposed_framework} presents the overall proposed framework and the methods used to address these challenges. Systematic experimentation and evaluation of the framework across different users, devices, iterations, models, labels in a federated learning setting is showcased in section \ref{section:exp_results}, while also proving feasibility of the same on resource-constrained devices (section \ref{section:on-device}). Finally, section \ref{section:conclusion} concludes the paper.

\section{Our Approach}
\label{section:our_approach}

In this section, we discuss in detail about the problem formulation of heterogeneity in labels and models, and our proposed framework to handle the same (showcased in Figure \ref{fig:block_diagram}).

\subsection{Problem Formulation}
\label{section:problem_formulation}

We assume the following scenario in federated learning. There are multiple local devices which can characterize different model architectures based on the end users. The incoming streaming real-world data in all the devices is non-IID in nature, wherein the distribution and data characteristics differ across devices. We hypothesize that the incoming data to the different devices also consist of heterogeneities in labels, with either unique or overlapped labels. We also have a public dataset with the label set consisting of all labels pre-defined -- this can be accessed by any device anytime, and acts as an initial template of the data and labels that can stream through, over different iterations. We repurpose this public dataset as the test set also, so that consistency is maintained while testing. To make different FL iterations independent from the public dataset, we do not involve the public dataset during federated learning (training) in the local models. The research problem here is to create a unified framework to handle heterogeneous models, labels and data distributions (with non-IID nature) in a federated learning setting.

\subsection{Proposed Framework}
\label{section:proposed_framework}

The proposed framework and methods to handle the heterogeneous labels and models in a federated learning setting is presented in Algorithm \ref{alg:proposed_method}.  There are three important steps in the proposed method.

\begin{algorithm}[ht]
\caption{Proposed Framework to handle heterogeneous labels and models in Federated Learning}

\textbf{Input} - Public Data set $\mathcal{D}_0\{x_0,y_0\}$, Private datasets $\mathcal{D}_m^i$, Total users $M$, Total iterations $I$, LabelSet for each user $l_m$
\vskip 0.1cm
\textbf{Output} - Trained Model scores $f_G^I$
\vskip 0.1cm
\textbf{Initialize} - $f_G^0 = \mathbf{0}$ (Global Model Scores)
\vskip 0.1cm
\textbf{for} $i = 1$ \textbf{to} $I$ \textbf{do}
\vskip 0.1cm
\hspace{5pt}\textbf{for} $m = 1$ \textbf{to} $M$ \textbf{do}
\vskip 0.1cm
\hspace{10pt}\textbf{Build}: Model $\mathcal{D}_m^i$ and predict $f_{\mathcal{D}_m^i}(x_0)$
\vskip 0.1cm
\hspace{10pt}\textbf{Local Update}: $f_{\mathcal{D}_m^i}(x_0) = f_G^I(x_0^{l_m)}+\alpha f_{\mathcal{D}_m^i}(x_0)$, where \\ 
\hspace{10pt}$f_G^I(x_0^{l_m)}$ are the global scores of only the set of labels $l_m$ \\
\hspace{10pt}with the $m^{th}$ user, and $\alpha = \frac{len(\mathcal{D}_m^i)}{len(\mathcal{D}_0)}$
\vskip 0.1cm
\hspace{10pt}\textbf{Global Update}: Update label wise,

\hspace{10pt}$f_G^{i+1} = \displaystyle \sum_{m=1}^{M}\beta_m f_{\mathcal{D}_m^i}(x_0)$, where,

\begin{align*}
    \beta = \begin{cases} 1 & \text{If labels are unique} \\
    \text{acc}(f_{\mathcal{D}_m^i}(x_0)) & \text{If labels overlap}
    \end{cases}
\end{align*}

\vskip 0.1cm
\hspace{5pt}\textbf{end for}
\vskip 0.1cm
\textbf{end for}
\label{alg:proposed_method}
\end{algorithm}

\subsubsection*{\textbf{Build:}} In this step, we build the model on the incoming data we have in each local user, i.e., local private data for the specific iteration. The users can choose their own model architecture which suits best for the data present in the iteration.
    
\subsubsection*{\textbf{Local Update:}} In this step, we update the averaged global model scores (on public data) for the $i^{th}$ iteration on the local private data. For the first iteration, we do not have any global scores and we initialize the scores to be zero in this case. For the rest of iterations, we have global averaged scores which we can use to update the local model scores according to Algorithm \ref{alg:proposed_method}. In the local update, we propose a \textit{weighted $\alpha$-update}, where $\alpha$ is the ratio between the size of current private dataset and the size of public dataset, and governs the contributions of the new and the old models across different federated learning iterations.
    
\subsubsection*{\textbf{Global Update:}} In this step, we first train the local model on the respective private datasets for that FL iteration. Further, we evaluate (test) this trained model on the public data, thereby obtaining the model scores on public data. We then perform the same operation across all the users and average them using the $\beta$ parameter, where $\beta$ governs the weightage given to overlapping labels across users using test accuracies of the corresponding labels on public data (as given in Algorithm \ref{alg:proposed_method}). This module gives the global averaged scores.

\begin{figure*}[ht]
  \centering
  \begin{subfigure}[b]{0.33\textwidth}
    \includegraphics[width=\linewidth]{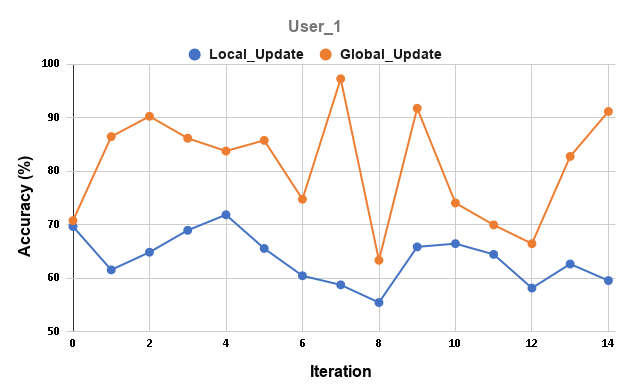}
    \caption{User\_1}
    \label{fig:user_1}
  \end{subfigure}%
  \begin{subfigure}[b]{0.33\textwidth}
    \includegraphics[width=\linewidth]{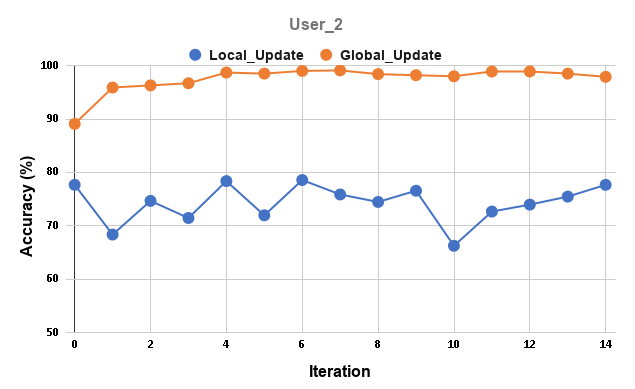}
    \caption{User\_2}
    \label{fig:user_2}
  \end{subfigure}%
  \begin{subfigure}[b]{0.33\textwidth}
    \includegraphics[width=\linewidth]{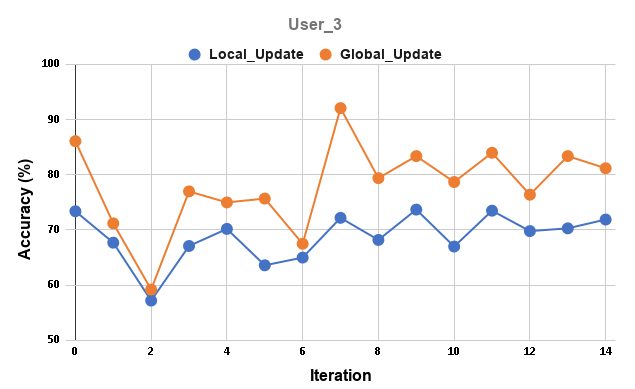}
    \caption{User\_3}
    \label{fig:user_3}
  \end{subfigure}%
    \centering
  \caption{Iterations vs Accuracies across all three users with Weighted Local Update with $\alpha$. \textit{Local\_Update} signifies the accuracy of each local updated model (after $i^{th}$ iteration) on Public Dataset. \textit{Global\_Update} signifies the accuracy of the corresponding global updated model (after $i^{th}$ iteration) on Public Dataset.}
  \label{fig:results}
\end{figure*}

\section{Experiments and Results}
\label{section:exp_results}

We simulate a Federated Learning scenario with multiple iterations of small chunks of incremental data incoming (details in Table \ref{table:models_labels_iterations}), across three different users to test our approach, and assume that the images arrive in real-time in the users' devices. We use the \textit{Animals-10 dataset} \cite{cite:animals_kaggle}, which consists of ten different animal categories (labels) extracted from Google Images. Now, we discuss the settings for label and model heterogeneities in our experiment.

\subsubsection*{\textbf{Label Heterogeneities:}} In our experiment, we consider only four labels -- \{\textit{Cat, Dog, Sheep, Elephant}\} from the dataset as shown in Table \ref{table:models_labels_iterations}. Also, we include the number of images considered per user per iteration (500 images per iteration). The labels in each local user can either be unique (present only in that single user) or overlapped across users (present in more one user). We split the four labels into three pairs of two labels each, for convenience of showcasing the advantage of overlapping labels in experimentation. We also create a non-IID environment across different federated learning iterations wherein, the image data across different iterations are split with disparities in both labels and distributions of data (\textit{Statistical Heterogeneities}).

\begin{table}[ht]
\caption{Details of Model Architectures (filters/units in each layer indicated within parentheses) changed across federated learning iterations and users.}
\label{table:changing_models}
\begin{tabular}{|c|c|}
\hline
\textbf{Iteration}    & \textbf{New Model Architecture}                                                                \\ \hline
User\_1 Iteration\_10 & \begin{tabular}[c]{@{}c@{}}3-Layer ANN\\ (16, 16, 32)\\ ReLU Activation\end{tabular}       \\ \hline
User\_1 Iteration\_14 & \begin{tabular}[c]{@{}c@{}}1-Layer CNN\\ (16)\\ Softmax Activation\end{tabular}              \\ \hline
User\_2 Iteration\_6  & \begin{tabular}[c]{@{}c@{}}3-Layer CNN\\ (16, 16, 32)\\ Softmax activation\end{tabular}    \\ \hline
User\_3 Iteration\_5  & \begin{tabular}[c]{@{}c@{}}4-Layer CNN\\ (8, 16, 16, 32)\\ Softmax activation\end{tabular} \\ \hline
\end{tabular}
\end{table}

\subsubsection*{\textbf{Model Heterogeneities:}} We choose three different model architectures for the three different local user device. This is clearly elucidated in Table \ref{table:models_labels_iterations}. To truly showcase near-real-time heterogeneity and model independence, we change the model across and within various FL iterations as shown in Table \ref{table:changing_models}.

Initially, we divide the images across different users according to the four labels. Each image is re-scaled to size 128*128. We create a Public Dataset - $D_0$ with 2000 images, with 500 images corresponding to each label. Next, we sample 500 images in every iterations as per the labels of the user and we use them for our problem (as shown in Table \ref{table:models_labels_iterations}). In total, we ran 15 iterations in this whole federated learning experiment, with each iteration running with early stopping (max 5 epochs). We track the loss using categorical cross-entropy loss function for multi-class classification, and use the Adam optimizer \cite{cite:adam_optim} to optimize the classification loss. We simulate all our experiments -- both federated learning and inference on a \textit{Raspberry Pi 2}.

\subsection{Discussion on Results}
\label{section:discussion_results}

Figure \ref{fig:results} represents the results across all three users on Animals Dataset. Also, from Table \ref{table:avg_animals}, we can clearly observe that the global updates -- which represent the accuracies of the global updated model (and averaged across all users' labels in the $i$th iteration governed by $\beta$), are higher for all three users than the accuracies of their respective local updates. For instance, from Figure \ref{fig:user_1}, we can infer that the corresponding accuracies of labels \{\textit{Cat, Dog}\} (User 1 labels) after global updates in each iteration are deterministically higher than their respective local updates by an average of $\sim$17.4\% across all iterations with weighted $\alpha$-update. Similarly for User 2, labels consisting of \{\textit{Dog, Sheep}\}, we observe an average accuracy increase of $\sim$23.2\% from local updates to the global updates, while for User 3 labels consisting of \{\textit{Sheep, Elephant}\}, we observe an average increase of $\sim$9.3\% from local updates to global updates.

\begin{table}[ht]
\caption{Average Accuracies (\%) of Local and Global Updates, and their respective Accuracy increase with \textit{Weighted $\alpha$-update}}
\label{table:avg_animals}
\begin{tabular}{|c|c|c|c|}
\hline
                 & \textbf{Local\_Update} & \textbf{Global\_Update} & \textbf{Acc. Increase} \\ \hline
\textbf{User\_1} & 63.66                  & 81.02                   & 17.36                      \\ \hline
\textbf{User\_2} & 74.3                   & 97.47                   & \textbf{23.17}             \\ \hline
\textbf{User\_3} & 68.72                  & 78.02                   & 9.3                        \\ \hline
\textbf{Average} & \textbf{68.89}         & \textbf{85.5}           & \textbf{16.61}              \\ \hline
\end{tabular}
\end{table}

We would like to particularly point out that the overlap in labels significantly contributed to highest increase in accuracies, owing to the fact that information gain (weighted global update) happens only for overlapping labels. This is vividly visible in User 2 (Figure \ref{fig:user_2}, whose labels are \{\textit{Dog, Sheep}\}), where in spite of an accuracy dip in local update at iteration 10, the global update at that iteration does not take a spike down which can be primarily attributed to the information gain from overlapped labels between User 1 and User 3 (in this case, \textit{Dog} and \textit{Sheep} respectively), thereby showcasing the robustness of overlapped label information gain in User 2. On the contrary, when we observe User 1 (Figure \ref{fig:user_1}), in spite of the accuracies of global updates being inherently better than local updates, when a dip in accuracies of local updates are observed at iteration 8 and 12, the accuracies of global updates at that iteration also spike down in a similar fashion. Similar trends of local and global accuracy trends like those observed in User 1 can also be observed in User 3 (Figure \ref{fig:user_3}). This clearly shows that when there are lesser overlapped labels (User 1 and User 3), the global model does not learn the label characteristics as much as when there are more overlapped labels -- the global updates are more robust in spite of spikes and dips in local updates with overlapped labels (User 2), thereby leading to higher average increase in accuracies (as observed in Table \ref{table:avg_animals}).

\begin{figure}[ht]
  \centering
  \includegraphics[width=\linewidth]{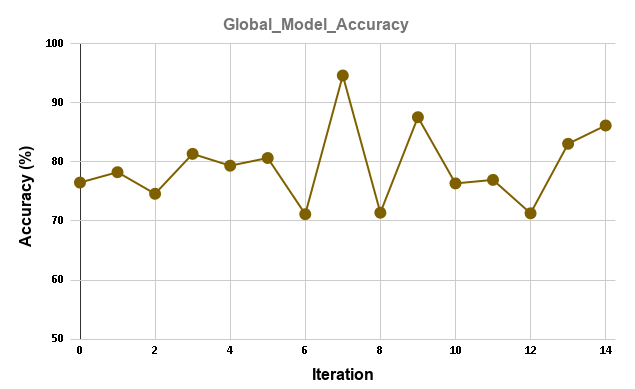}
  \caption{Iterations vs Final Global Average Accuracies (\%) with \textit{Weighted $\alpha$-update}}
  \label{fig:global_results}
  \Description{Global Model Accuracy Graph}
\end{figure}

An overall average increase of $\sim$16.7\% deterministic accuracy increase (not relative) is observed with our proposed framework across all the different users, which can be calculated from the global model updates. The overall global model accuracies alone (which is different from global update accuracies observed in Figure \ref{fig:results}) after each iteration are also elucidated in Figure \ref{fig:global_results}. We propose that, with our current framework, communication (transfer) of just the model scores of respective labels between clients (local devices) and the central cloud is enough, without necessitating transfer of the whole model weights, which significantly reduces latency and memory overheads. We also point out that we tried local model distillation similar to the work done in FedMD \cite{cite:FedMD} for handling label and model heterogeneities, however, we felt using only score consensus from the model without distillation yielded far lesser on-device computation times.

\subsection{On-Device Performance}
\label{section:on-device}

We observe the on-device performance of our proposed framework, which is experimented on a Raspberry Pi 2. We choose this single-board computing platform since it has similar hardware and software (HW/SW) specifications with that of predominant contemporary IoT/edge/mobile devices. The computation times taken for execution of on-device federated learning and inference are reported in Table \ref{table:timeOnPi}. This clearly shows the feasibility of our proposed system on embedded devices.

\begin{table}[ht]
\caption{Time taken for Execution}
\label{table:timeOnPi}
\begin{tabular}{|c|c|}
\hline
\textbf{Process}                                                                     & \textbf{Computation Time} \\ \hline
\begin{tabular}[c]{@{}c@{}}Training time per epoch\\ in an FL iteration ($i$)\end{tabular} & $\sim$1.8 sec                                           \\ \hline
Inference time      & $\sim$15 ms                       \\ \hline
\end{tabular}
\end{table}

\section{Conclusion}
\label{section:conclusion}

This paper presents a unified framework for flexibly handling heterogeneous labels and model architectures in federated learning in a non-IID fashion. By leveraging transfer learning along with simple scenario changes in the federated learning setting, we propose a framework with $\alpha$-update aggregation in local models, and we are able to leverage the effectiveness of global model updates across all devices and obtain higher efficiencies. Moreover, overlapping labels are found to make our framework robust, and also helps in effective accuracy increase.  We empirically showcase the successful feasibility of our framework on resource-constrained devices for federated learning/training and inference across different iterations on the Animals-10 dataset. We expect a good amount of research focus hereon in developing statistical, model and label based heterogeneities.

\bibliographystyle{acm}
\bibliography{References}

\begin{thebibliography}{10}

\bibitem{cite:animals_kaggle}
{\sc Alessio, C.}
\newblock {\em Animals-10 Dataset}, 2019.
\newblock \url{https://www.kaggle.com/alessiocorrado99/animals10}.

\bibitem{cite:federated_learning}
{\sc Bonawitz, K., Eichner, H., Grieskamp, W., Huba, D., Ingerman, A., Ivanov,
  V., Kiddon, C., Konecny, J., Mazzocchi, S., McMahan, H.~B., Overveldt, T.~V.,
  Petrou, D., Ramage, D., and Roselander, J.}
\newblock Towards federated learning at scale: System design.
\newblock In {\em SysML 2019\/} (2019).

\bibitem{cite:personalized_fed}
{\sc Deng, Y., Kamani, M.~M., and Mahdavi, M.}
\newblock Adaptive personalized federated learning.
\newblock {\em arXiv preprint arXiv:2003.13461\/} (2020).

\bibitem{cite:federated_heterogeneous}
{\sc Ghosh, A., Hong, J., Yin, D., and Ramchandran, K.}
\newblock Robust federated learning in a heterogeneous environment.
\newblock {\em arXiv preprint arXiv:1906.06629\/} (2019).

\bibitem{cite:knowledge_distillation}
{\sc Hinton, G., Vinyals, O., and Dean, J.}
\newblock Distilling the knowledge in a neural network.
\newblock In {\em NIPS Deep Learning and Representation Learning Workshop\/}
  (2015).

\bibitem{cite:fed_distill_aug}
{\sc Jeong, E., Oh, S., Kim, H., Park, J., Bennis, M., and Kim, S.-L.}
\newblock Communication-efficient on-device machine learning: Federated
  distillation and augmentation under non-iid private data.
\newblock {\em arXiv preprint arXiv:1811.11479\/} (2018).

\bibitem{cite:advances}
{\sc Kairouz, P., McMahan, H.~B., Avent, B., Bellet, A., Bennis, M., Bhagoji,
  A.~N., Bonawitz, K., Charles, Z., Cormode, G., Cummings, R., et~al.}
\newblock Advances and open problems in federated learning.
\newblock {\em arXiv preprint arXiv:1912.04977\/} (2019).

\bibitem{cite:adam_optim}
{\sc Kingma, D.~P., and Ba, J.}
\newblock Adam: A method for stochastic optimization.
\newblock {\em arXiv preprint arXiv:1412.6980\/} (2014).

\bibitem{cite:federated_opt}
{\sc Kone{\v{c}}n{\`y}, J., McMahan, H.~B., Ramage, D., and Richt{\'a}rik, P.}
\newblock Federated optimization: Distributed machine learning for on-device
  intelligence.
\newblock {\em arXiv preprint arXiv:1610.02527\/} (2016).

\bibitem{cite:federated_learning_strategies}
{\sc Kone{\v{c}}n{\`y}, J., McMahan, H.~B., Yu, F.~X., Richt{\'a}rik, P.,
  Suresh, A.~T., and Bacon, D.}
\newblock Federated learning: Strategies for improving communication
  efficiency.
\newblock In {\em NIPS Workshop on Private Multi-Party Machine Learning\/}
  (2016).

\bibitem{cite:FedMD}
{\sc Li, D., and Wang, J.}
\newblock Fedmd: Heterogenous federated learning via model distillation.
\newblock {\em arXiv preprint arXiv:1910.03581\/} (2019).

\bibitem{cite:challenges_methods}
{\sc Li, T., Sahu, A.~K., Talwalkar, A., and Smith, V.}
\newblock Federated learning: Challenges, methods, and future directions.
\newblock {\em IEEE Signal Processing Magazine 37}, 3 (2020), 50--60.

\bibitem{cite:fedprox_heterogeneity}
{\sc Li, T., Sahu, A.~K., Zaheer, M., Sanjabi, M., Talwalkar, A., and Smith,
  V.}
\newblock Federated optimization in heterogeneous networks.
\newblock {\em Proceedings of Machine Learning and Systems 2020\/} (2020),
  429--450.

\bibitem{cite:fair_resource_allocation}
{\sc Li, T., Sanjabi, M., Beirami, A., and Smith, V.}
\newblock Fair resource allocation in federated learning.
\newblock In {\em 8th International Conference on Learning Representations
  (ICLR)\/} (2020).

\bibitem{cite:convergence_non_iid}
{\sc Li, X., Huang, K., Yang, W., Wang, S., and Zhang, Z.}
\newblock On the convergence of fedavg on non-iid data.
\newblock In {\em 8th International Conference on Learning Representations
  (ICLR)\/} (2020).

\bibitem{cite:federated_mobile_survey}
{\sc Lim, W. Y.~B., Luong, N.~C., Hoang, D.~T., Jiao, Y., Liang, Y.-C., Yang,
  Q., Niyato, D., and Miao, C.}
\newblock Federated learning in mobile edge networks: A comprehensive survey.
\newblock {\em IEEE Communications Surveys \& Tutorials\/} (2020).

\bibitem{cite:fedavg}
{\sc McMahan, H.~B., Moore, E., Ramage, D., Hampson, S., and Arcas, B. A.~y.}
\newblock Communication-efficient learning of deep networks from decentralized
  data.
\newblock In {\em Proceedings of the 20th International Conference on
  Artificial Intelligence and Statistics\/} (2017), vol.~54, pp.~1273--1282.

\bibitem{cite:agnostic_federated}
{\sc Mohri, M., Sivek, G., and Suresh, A.~T.}
\newblock Agnostic federated learning.
\newblock In {\em Proceedings of the 36th International Conference on Machine
  Learning\/} (2019), vol.~97, pp.~4615--4625.

\bibitem{cite:client_selection}
{\sc Nishio, T., and Yonetani, R.}
\newblock Client selection for federated learning with heterogeneous resources
  in mobile edge.
\newblock In {\em IEEE International Conference on Communications (ICC)\/}
  (2019), pp.~1--7.

\bibitem{cite:federated_non_iid}
{\sc Zhao, Y., Li, M., Lai, L., Suda, N., Civin, D., and Chandra, V.}
\newblock Federated learning with non-iid data.
\newblock {\em arXiv preprint arXiv:1806.00582\/} (2018).

\end{thebibliography}

\end{document}